# *A Deep Generative Artificial Intelligence system to decipher species coexistence patterns*

Hirn, J.[1], García, J.E[1], Montesinos-Navarro, A.[2], Sánchez-Martín, R[2], Sanz[1,3], V. & Verdú, M[2*]

Authors affiliation:

1. Instituto de Física Corpuscular (IFIC, Universidad de Valencia-CSIC), E-46980 Valencia, Spain.
2. Centro de Investigaciones Sobre Desertificación (CIDE, CSIC-Universidad de Valencia-Generalitat Valenciana), E-46113, Valencia Spain.
3. Department of Physics and Astronomy, University of Sussex, Brighton BN1 9QH, UK.

*Corresponding Author

Running headline
*Artificial intelligence and species coexistence*

**Abstract**

1. Deciphering coexistence patterns is a current challenge to understanding diversity maintenance, especially in rich communities where the complexity of these patterns is magnified through indirect interactions that prevent their approximation with classical experimental approaches.
2. We explore cutting-edge Machine Learning techniques called Generative Artificial Intelligence (GenAI) to decipher species coexistence patterns in vegetation patches, training generative adversarial networks (GAN) and variational AutoEncoders (VAE) that are then used to unravel some of the mechanisms behind community assemblage.
3. The GAN accurately reproduces the species composition of real patches as well as the affinity of plant species to different soil types, and the VAE also reaches a high level of accuracy, above 99%. Using the artificially generated patches, we found that high order interactions tend to suppress the positive effects of low order interactions. Finally, by reconstructing successional trajectories we could identify the pioneer species with larger potential to generate a high diversity of distinct patches in terms of species composition.
4. Understanding the complexity of species coexistence patterns in diverse ecological communities requires new approaches beyond heuristic rules. Generative Artificial Intelligence can be a powerful tool to this end as it allows to overcome the inherent dimensionality of this challenge.

# 1 INTRODUCTION

Understanding how species coexist has always been a central problem in ecology, given that it is at the core of diversity maintenance (Chesson, 2000). The complexity of coexistence patterns is magnified in diverse communities where coexistence is not only a signal of paired interactions but also of indirect interactions (Strauss, 1991). Thus, the probability that two species coexist depends on the presence of a third, fourth, fifth, or n-th species. Experimental approaches have been commonly used to explore coexistence patterns, although their focus on pairs or on a few sets of species lack realism (van Kleunen, 2014). This is unavoidable, as the number of indirect interactions increases exponentially with the number of species considered, precluding the experimental quantification of all of them. Therefore, other tools are currently needed to assess coexistence patterns in rich ecological communities where tens, hundreds, or even thousands of species coexist.

Machine learning is able to detect complex patterns beyond heuristic rules and traditional statistics. Neural Networks, a form of Machine Learning, can detect intricate patterns produced by high order interactions as those produced among genes in regulatory networks (Libbrecht & Noble 2015) or genetic, clinical and histological variables in cancer diagnosis (Kourou et al 2015). In ecology, deep learning has often been used to assist researchers in processing large datasets produced by automatic monitoring of populations and ecosystems by applying deep neural networks (Christin et al 2019). However, this methodology can be applied much more broadly and the road is paved to study high dimensional problems related to ecological interactions (Desjardins‑Proulx et al., 2017).

Species coexistence in nature follows complex, unknown patterns that Machine Learning, with its high degree of expressivity, should be able to capture. Here, we explore the use of a set of cutting-edge Machine Learning techniques called Generative Artificial Intelligence (GenAI) (Ruthotto and Haber, 2021) to decipher species coexistence patterns that could later be used to unravel the mechanisms behind community assemblage. The word Generative indicates the ability of these techniques to create new, unseen situations from a limited data set of examples. Among the Generative AI methods, we choose the two most powerful ones: Generative Adversarial Networks (GAN) and Variational AutoEncoders (VAE), both with their own strengths and weaknesses. On the one hand, GANs consist of two models that are simultaneously trained, so that a generative model G captures the distribution of the data, and a discriminative model D estimates the probability that a sample comes from the training data rather than from G (Fig 1 left). The training for G maximizes the likelihood that the discriminative model makes a mistake (Goodfellow et al., 2014). On the other hand, VAEs are generative machine learning models that combine a pair of neural networks that aim to first compress and then mirror the input data given a set of latent coordinates (Kingma and Welling 2013) (Fig 1 right). VAEs incorporate nonlinear relationships and allow users to define the dimensionality of the latent space. Besides, the loss function for a VAE can be used to assess the difference between a reference distribution (a prior in the latent space) and a population's distribution viewed in the same latent space and (Battey et al., 2021). GANs have shown great ability to generate realistic avatars, whereas the more complex structure of VAEs allows us to ask deeper questions, at the level of the latent space (Ruthotto and Haber 2021).

In this study, we introduce the application of generative artificial intelligence to assess plant coexistence patterns and illustrate its potential in a facilitation-driven community where plants tend to grow together, forming vegetation patches (Montesinos-Navarro et al., 2019). In facilitation-driven patches, indirect interactions occur, and the coexistence of species within a patch strongly depends on the composition of the neighborhood (Castillo et al., 2010; Schöb et al., 2013). We characterize plant species composition of vegetation patches, which could be used to estimate the probability of species co-occurrence across them. With an unlimited number of patches sampled, this probability would become a theoretical distribution of all species co-occurrence, and thus a manifestation of the underlying rules dictating the patch composition. Hence, we develop a Machine Learning method able to model a continuous probability distribution from a finite set of observations. It predicts the most likely compositions of patches, guided by the underlying coexistence rules, which cannot be represented in simple ways. We trained two Generative Artificial Intelligence systems (GAN and VAE) to generate fake but likely compositions of patches (hereafter fake patches), and validate them with ecological questions of increasing complexity. First, we assess whether the fake patches mirrored 1) the relative abundance of patches with a given species composition and 2) the plant species affinity to different soil types. Then, we used the GAN to 3) assess the relative contribution of direct and indirect interactions in determining the probability of species co-occurrence in vegetation patches, and the VAE to 4) forecast the amount of potential species compositions in fake patches following the succession triggered by a pioneer species. Finally, we provide guidelines to construct personalized GenAI models and the code to run them.

## 2 MATERIALS AND METHODS

### *2.1. Input data*

The species composition of 5153 vegetation patches was characterized in 4 dryland plant communities (hereafter sites) situated within a radius of 20 Km in Alicante (southeast Spain). Within each site, the vegetation patches were distributed in two adjacent soil types (hereafter gypsum and limestone subsites) located less than 10 m apart, minimizing the potential effect of dispersal limitation of species between subsites. The sampling design comprised 80 plots (150x150cm) randomly distributed in each subsite, except one subsite with 79 plots. Inside each plot, we identified and registered all the species present in each vegetation patch. A patch is composed of at least two individuals of different species surrounded by bare ground, with a mean surface area of $512 \pm 982$ cm$^2$.

Vegetation patches are expressed as arrays of presence/absence of plant species in an ordered list of *n* species where 1 would denote presence and 0 absence (e.g. *x*= (1,0,0,0,1,...,0)).. There are as many arrays as vegetation patches sampled (*N*=5153), so that our database is then a list of vectors {$x_1, x_2, x_3, ..., x_N$} in a $\Re^n$ space, which includes a number of species (we will choose the most abundant species for illustrative purposes), and also the soil type (1=gypsum or 0=limestone) in which that vegetation patch was observed. These vectors will be the input given to the GenAI network to learn co-occurrence patterns among species.

### *2.2 Generative Artificial Intelligence Systems*

In the following we describe the two techniques employed in this study, GAN and VAE.

#### *2.2.1. Generative adversarial networks (GAN)*

We trained three different GANs, denoted by GAN*n*, with different dimensionalities *n* indicating the number of plant species considered in each of them (most abundant), and soil type: GAN8, GAN16, and GAN32. During the training, the GAN takes each real patch and creates a fake patch. At the beginning, the fake patches are very different from the real ones, but the GAN trains adversarially until the fake and the real patches are indistinguishable from each other. At that point, the GAN has *learned the rules of the game,* namely it can not just produce the initial real patches, but also any new fake patches which represent suitable possibilities. This is the generative feature of GANs, the ability to generate an infinite number of fake patches that were not found in the original dataset but reflect likely species' composition based on the plant community assembly rules acquired through training.

We use the *Python* library *fastai 2.1.5* to train a basic Wasserstein GAN with 10 dimensions in the input space, one extra layer in the generator and one in the critic, using ReLU activation functions with a negative slope of 0.2. The GAN is trained with RMSProp optimizer and $2 \times 10^{-4}$ learning rate on 2D square images each representing the composition of a single patch. We construct these 2D images by taking as one direction the pattern of zeroes and ones describing the absence and presence of a given species or soil type in that patch, and repeating that pattern along a second dimension to form a square. After training for 2000 epochs, we produce 300,000 fake patches by feeding the GAN a 10-dimensional Gaussian noise. The GAN outputs 2D images which can be translated into absence/presence of species by removing the edges along the repeated dimension of the image, averaging along the remainder of that dimension, and finally using a threshold to discretize the output.

To estimate the systematic error of our procedure, we have performed the training procedure 14 times using the same number of epochs, but with independent, random initialization each time. The error bars in our figures depict two standard deviations around the mean, all computed over these 14 GAN runs. Since the GANs are trained on real patches with at least two species of plants (in order to avoid training on the numerous patches that contain only a single species), we also reject fake patches containing fewer than two species (about 7.5% of the fake patches produced by the GAN).

### 2.2.2. Variational AutoEncoder (VAE)

We explore the ability of VAE to learn subtle species interactions in a space of a large dimensionality. The VAE learns by looping around an encoder and decoder which transforms the real data into fake data. Our VAE takes as an input a rectangular grayscale 2D image. We extend the 1D line of zeros and ones representing the absence/presence of the 8, 16 or 32 most common species or soil type into a second dimension by repeating the line 8 times.

For the case of VAE8 we build a convolution VAE using the *Python* library *Keras 2.3.1* and three 2x2 convolution layers with stride 2, with successive numbers of filters 128, 256 and 512, then reduce this to fit a 128-dimensional latent space. The GAN is trained with Adam optimizer and $1 \times 10^{-4}$ learning rate. We obtain 99.90% accuracy on a pixel-by-pixel level for our best model. To turn our rectangular monochrome images back into binary presence/absence of plants or soil type, we remove the edges along the repeated dimension of the image, average along the remainder of that dimension, and finally use a threshold to discretize the output. This translation of the output patch agrees with the input one for 99.19% of the patches.

### 2.3 Ecological Validation: patch species composition and plant soil affinity

In order to validate whether the GAN can produce species co-occurrence patterns similar to those observed in the system, we focus on two features: The relative abundance of the different species compositions of patches, and the affinity of certain species to a given soil type. We use each GAN to generate 300K fake patches, and then compare the features of these fakes with those of the real patches characterized in the field. Firstly, we quantify the relative abundance of patches with unique species compositions, and compare the relative abundances between real and fake patches. Secondly, we calculate the probability that a given species is found in gypsum vs limestone soil, estimating its affinity for gypsum (= absence of limestone). Finally, we compared whether the correlation estimated for each plant species is similar when it is based on the real and the fake patches.

### 2.4 Contribution of direct and indirect interactions to species coexistence

Direct and indirect interactions among species can be quantified using conditional probabilities. To study direct pairwise interactions one can compute the probability $P(A|B)$ = probability that species A is present when B is present. In Figure 4A, we represent this conditional probability $P(A|B)$ as a function of the relative abundance of species A in the GAN8 analysis, $P(A)$. Different colors represent different choices of species A, and the circle sizes are related to the relative amount of a particular combination AB. The values of $P(A)$ are found in the range of 25% to 45%, as GAN8 is trained with the most abundant species.

If there were no interactions between A and B, their probabilities should be independent P(A) = P(A|B), a situation which would follow the dashed trend line. Instead we observe that for a given species A, circles of the same color along the vertical axis, P(A|B) lies outside that line, which corresponds to sizeable interactions between A and B. Points above the dashed line indicate enhanced coexistence, and below depressed coexistence. Note that the quantity P(A|B) is not symmetric, i.e. P(A|B) - P(B|A) is not necessarily zero, as situations when A or B are pioneers may be different. For example for A=*Fumana thymifolia* and B=*Brachypodim retusum*, P(A|B)=0.5 and P(B|A)=0.4.

To study indirect interactions we can compute conditional probabilities involving three or more plants. In the two lower panels of Figure 4 we represent $3^{th}$ and $4^{th}$ order (indirect) interactions via conditional probabilities of presence of species A when BC and D are already present in the patch. In Figure 4B, points above the diagonal P(A|B) = P(A|BC) indicate that, in general terms, the presence of a third species enhances the co-occurrence of pairs, and points below imply that the third species suppresses the co-occurrence of pairs. The analysis can be carried over to higher-order interactions, as shown in Figure 4C, but paying the price of a lower probability represented by small circles.

**2.5 Forecasting the final composition of patches triggered by a pioneer species**

Once we trained the VAES as described in point 2.2 above, we started by considering a patch with a single pioneer, and then added to this patch a level of random noise. By processing this input through the VAE, we obtain a number of possible configurations with other species, whose abundance gives us information on the compatibility of each configuration. Moreover, we can exploit the VAE's ability to handle higher dimensionality and its abstract representation in terms of the latent space.

# 3 RESULTS

*3.1 Ecological Validation: patch species composition and plant soil affinity*

The patches generated by the GAN8 and VAE models do indeed reproduce the species composition of real patches, but also extend to produce new unseen but likely possibilities (Fig. 2). In particular, the fake patches generated by the GAN reproduce a similar abundance of patches with a unique composition compared to the patches found in the field (Fig 2a). Furthermore, the GAN is also able to produce new types of configurations beyond those used to train it (Fig 2b). While the real data (blue line in Fig 2b) shows a plateau in the number of patches with unique species composition due to the limited amount of field observations, the GAN results (red line in Fig 2b) can exceed that amount, showing its ability to generate new possible species composition in fake patches. Note that we find similar levels of capacity for learning for GAN8, GAN16 and GAN32, despite increasing dimensionality, when we compare abundances and interaction distributions between the real and fake generated patches. Yet for visualization we show results with the GAN8 model. A detailed comparison between the real data and the GAN results can be found on Datapane (https://datapane.com/u/johannes/reports/gan/ )**,** and the database and codes can be found in Github **(**https://github.com/jegarcian/AI4Ecology **)**

From an ecological perspective, the GAN accurately reproduces the plant species affinity to different types of soils (Fig. 3). The prediction was good enough for the whole range of plant species' soil affinities, including species with high, medium or low affinity to the two different types of soils.

*3.2 Contribution of direct and indirect interactions to species coexistence*

Direct interactions precluding species coexistence (dots below the diagonal in Fig 4a) were much more frequent than those promoting coexistence (dots below the diagonal in Fig 4a; t=-17.01; p<0.001; n=839), suggesting that most of the pairs of species seldom co-occur, either because species tend to live in different habitats (e.g. soil types) and/or to exclude competitively each other. Pairs of species with low probability to coexist were not affected by the presence of a third species but coexistence of those with high probability to coexist tended to be suppressed in the presence of a third species. Similarly, the effect of a fourth species reduced the positive effects of the third species, as we exemplify below.

For example, let us focus on the purple vertical set of points at x=0.5 in Fig 4b, which correspond to species A=*Fumana thymifolia*. From Figure 4a we know that this species is present in about 40% of the patches. All these points in Fig 4b around x=0.5 correspond to the co-existence with B=*Stipa tenacissima* or B=*Brachypodium retusum*, which boost the presence of *F. thymifolia* from 40% to 50%. But when another, third species C appears, the presence of *F. thymifolia* swings again in a wide range from a highly suppressed 10% (lower points at x=0.5) due to the presence of *Teucrium libanitis* or *Helianthemum squamatum,* to enhanced to 60% (higher points at x=0.5) due to the presence of *Stipa tenacissima*.

With the help of the GAN, we can go further than interactions among three. Fig 4c shows the distribution of indirect interactions among four species. We observe that the probability of finding a given species in a patch with three others rather than two is usually smaller. Indeed we expect that adding a fourth or fifth would reduce the chances of finding the species under consideration in most cases, as we have fewer and fewer patches with a large number of species. However, there are still many outliers indicating strong $4^{th}$ order interactions. For example, let us focus on the light green dots in Fig 4c, corresponding to A=*Helianthemum syriacum*. From Fig 4a we know that direct interactions do suppress the presence of this species, which on its own appears 42% of the time, but in lower frequencies when another species is present. From Fig 4b, we see that triple interactions do not overcome this suppression, with P(A|BC) always below 40%. But then in Fig 4c we see how the presence of yet another species, a $4^{th}$ order interaction, can change this trend, with a set of the light green combinations found above 40%. In particular, the $4^{th}$ order interactions resulting in an enhancement to 55% of the *Helianthemum syriacum* abundance are due to the co-occurrence with *Stipa tenacissima*, *Helianthemum squamatum* and *Fumana thymifolia*.

### 3.3. Forecasting the final composition of patches triggered by a pioneer species

We also train VAEs with 8, 16 and 32 species. After training the VAE reaches a high level of accuracy, above 99%. This accuracy means that the VAE is able to produce fake patches which strongly resemble the original patches. In particular, if we input a real patch composition, the VAE transformed patch would be identical to the input configuration in 99% of the cases.

We can exploit the VAE ability to represent the probability distribution in its latent space by, for example, inputting a pioneer species into this space and observing how the VAE generated probability distributions of generated patches with this pioneer, and thus select the best pioneer species. The results of VAE8 are shown in Figure 5, where we represent the distribution of unique patches generated by a single pioneer species introduced into the VAE's latent space. On the top of the plot, we observe that both *Helianthemum squamatum* and *Teucrium libanitis* as pioneer species produce a few independent types of patches with high probability (20% to 40% each) and seldom any other, quickly saturating close to 100% after about 10 unique patches. On the other hand, using *Fumana ericoides* and *Helianthemum syriacum* as pioneer species produces a wide range of distinct patches, each with a low probability (5% or less). These last two species therefore seem to encourage a wider biodiversity.

# 4 DISCUSSION

Species do interact in complex ways, with non-negligible indirect interactions leading to high boosting effects (Bairey et al. 2016). Therefore a simple set of rules involving two species would not capture the whole set of patterns emerging in a community. This complexity and the inherent dimensionality of this problem motivates the search for a new approach to describe coexistence patterns, beyond heuristic rules. Here we show that unsupervised machine learning methods based on generative artificial intelligence correctly predict a range of characteristics related to species coexistence. Just feeding the models with the species composition of 5153 vegetation patches in gypsum and limestone soils, we obtained correct predictions for i) the relative abundance of patches with different species composition; ii) the soil affinity of plant species, iii) the role of indirect interactions of third and fourth order in the coexistence of pairs of species. Furthermore, based on the association rules learnt, the model is able to predict the species composition of patches not registered in the field. This predictive power allows us to test a bunch of other ecological questions. For example, what would be the ecological succession given the colonization of a particular pioneer species.

In the context of species coexistence, the ability of Generative Artificial Intelligence to identify interactions of high order is specially relevant. Here we have identified third and fourth order interactions as an application with our dataset, but extensions to fifth and sixth order would be possible with a larger dataset.

The salient picture of these analyses is that a high order interaction tends to buffer the positive effects of the immediately lower-order interaction. For example, third order interactions tend to promote exclusion between pairs of species that tend to coexist. Similarly, fourth order interactions may depress the positive effects of coexistence produced by third order interactions. The magnitude of these depressing effects decrease with the order of the interaction, i.e. the effect of $4^{th}$ order interaction is smaller than that of $3^{rd}$ order interaction. Although this is the general trend there are also indirect interactions that positively promote coexistence (points above the diagonal in Figure 4) or that have no effect on it (points close to the diagonal in Figure 4). The final outcome of the high-order interaction is strongly dependent on the identity of the third (or fourth) species involved.

To elucidate the mechanism behind these patterns, future research could include phenotypic, phylogenetic or other relevant information to supervise the learning process and check whether the accuracy is improved. In our case, extra information seems unnecessary as the unsupervised VAE led to a very high accuracy, with 99% of the avatars strongly resembling the observed patches in the field. Thus, any phylogenetic or phenotypic pattern responsible for the species association has been learnt by the VAE and therefore we could ask it directly. For example, we could check whether the probability of two species with a low affinity to a particular stressful type of soil increases with the presence of a soil-specialist plant species, or whether the probability of exclusion between two closely-related species decreases with the presence of a third closely-related species. More generally, the predictions of the VAE could then help guide the efforts of enriching the ecosystem by introducing new species into it.

To get correct answers, the researcher should follow several steps. First, the field data should be collected for the particular question under scrutiny, minimizing the systematic disarticulation between data accumulation in public databases and knowledge production in our study (Devictor & Bensaude-Vincent, 2016). Second, sampling size should be enough to capture replicates of the interactions of order $n^{th}$. As a general rule, for *n* species, the number of interactions of order *k* will be $n!/(n-k)!$. For example, if we are interested in $3^{rd}$ order interactions in a community with 20 species, we will have $20!/(20-3)! = 6840$ possible combinations that should be replicated. Third, proper training and validation is needed to decide if the model is good enough to be used (see Christin et al. 2021). To further facilitate the application of this method to other datasets, the code used in this study has been made accessible in a repository (https://github.com/jegarcian/AI4Ecology) and detailed steps added to the documentation. Steps include:

pre-processing of the collected data, GAN/VAE training and data analysis, each of them has their own *notebook* which runs almost independently.

**ACKNOWLEDGEMENTS**
The authors thank the Yesaires team for making the fieldwork of quantification of species gypsum affinity possible. RSM was supported by the Ministry of Science and Innovations (FPU grant FPU17/00629). Financial support was provided by the Spanish Ministry of Science, Innovation, and Universities (RTI2018-099672-J-I00).

**AUTHOR´S CONTRIBUTIONS**
MV and AMN conceived the idea, VS, JH and JEG designed the methodology; RSM and AMN collected the data; VS, JH and JEG analysed the data; MV and VS led the writing of the manuscript. All authors contributed critically to the drafts and gave final approval for publication.

**Data Availability**

Data can be found at **datapane** (https://datapane.com/u/johannes/reports/gan/ )**,** and the code to run the analyses can be found at **github (**https://github.com/jegarcian/AI4Ecology **)**

**Figures**
**Figure 1.**

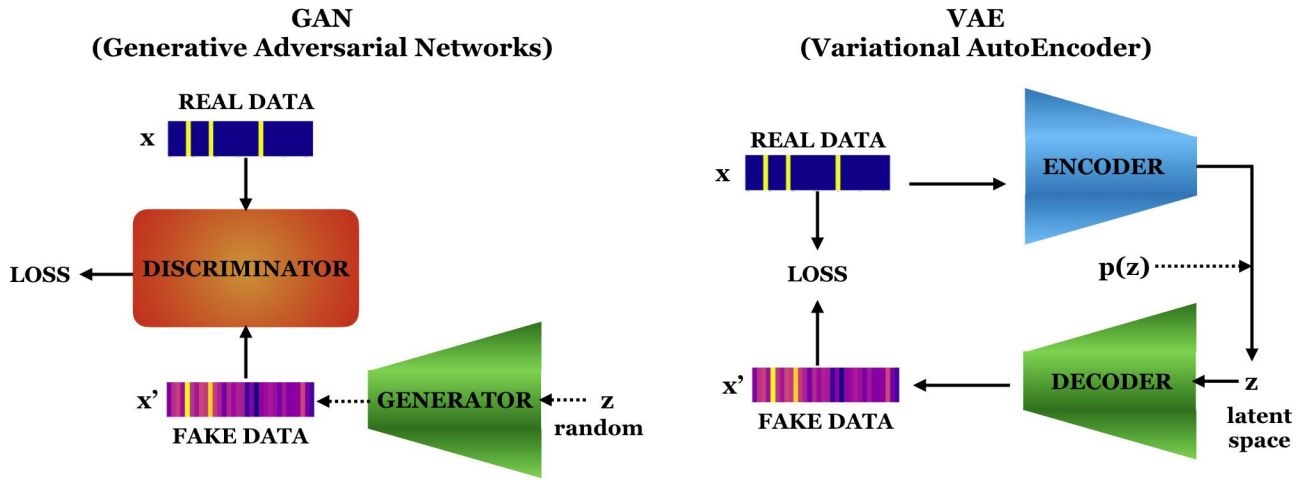

**Figure 1.** Schematic description of the GAN and VAE architectures. Real data is represented with the label x, and the generated data by x', whereas z denotes an external variable randomly generated. Generator, encoder and decoder are made of layers of artificial neurons.

a)

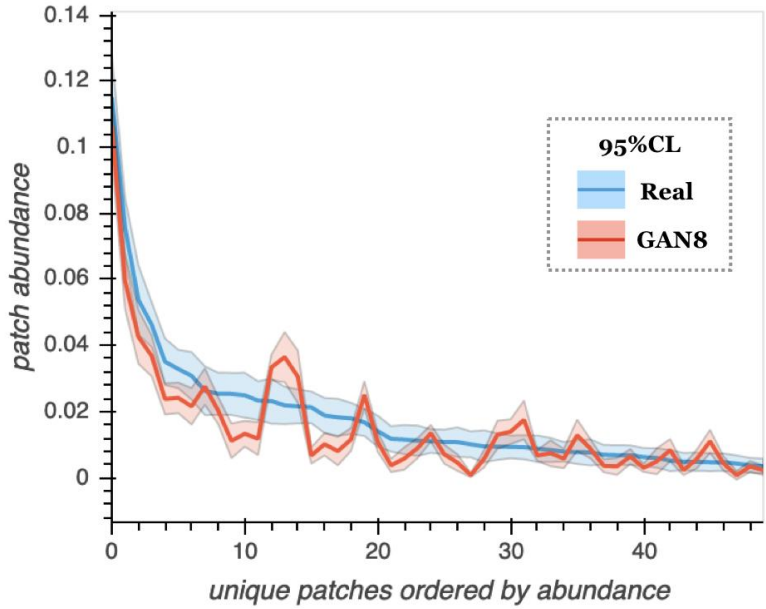

b)

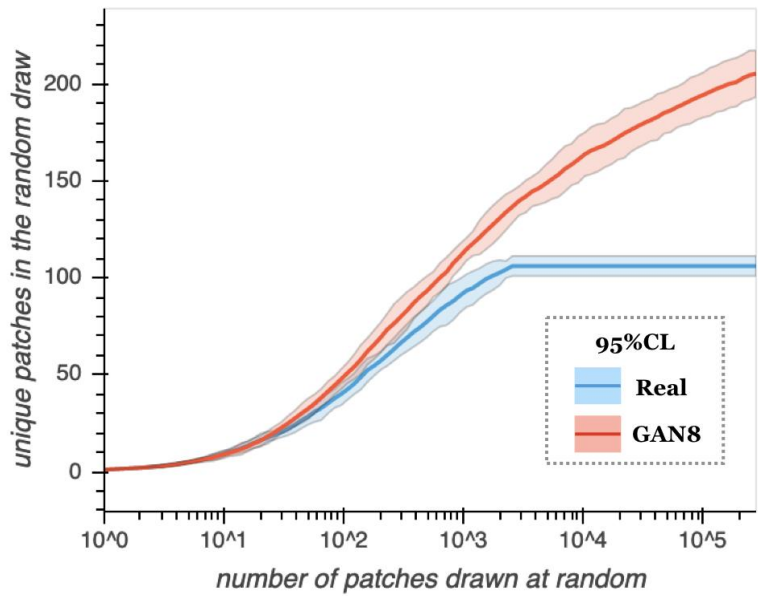

**Figure 2.** a) Comparison of the real abundances of patches with unique species composition and the abundances predicted with the patches generated by the GAN8 model. b) Number of patches with unique species composition in random samples of increasing size for both real and GAN8 patches. Abundances are shown with their 95% CL ranges.

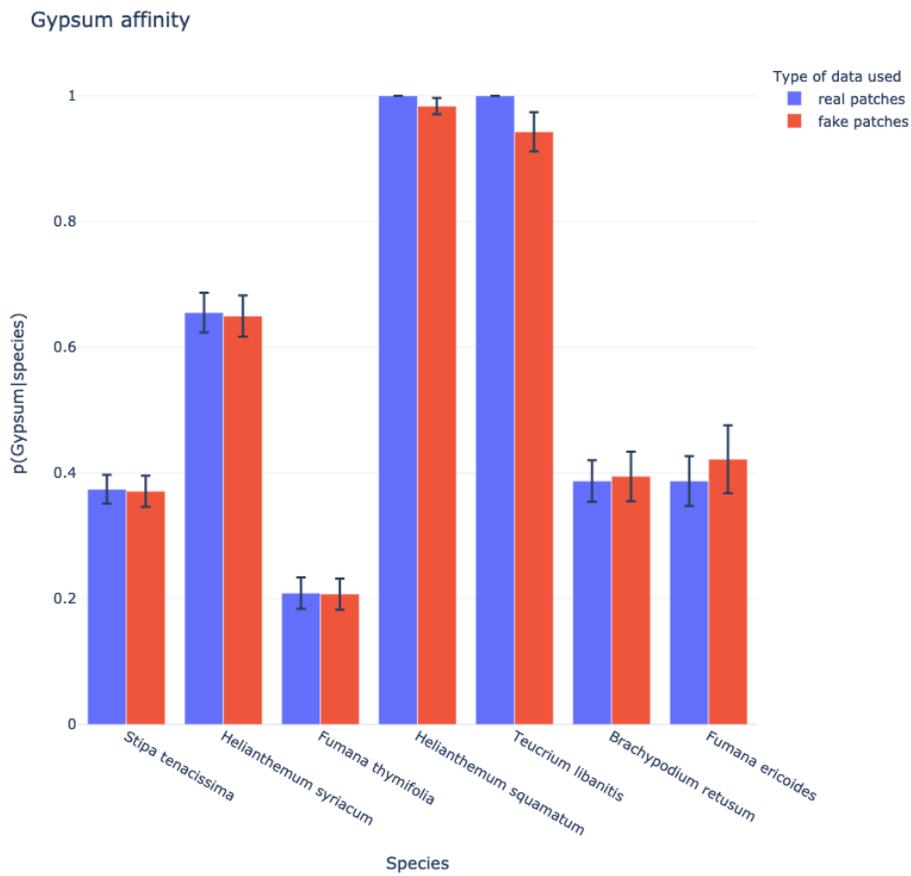

**Figure 3** Gypsum affinity in the observed patches and the GAN generated ones for the most abundant species. The error bars correspond to 95% CL due to statistical errors (real) and systematic errors (fake).

**a)**

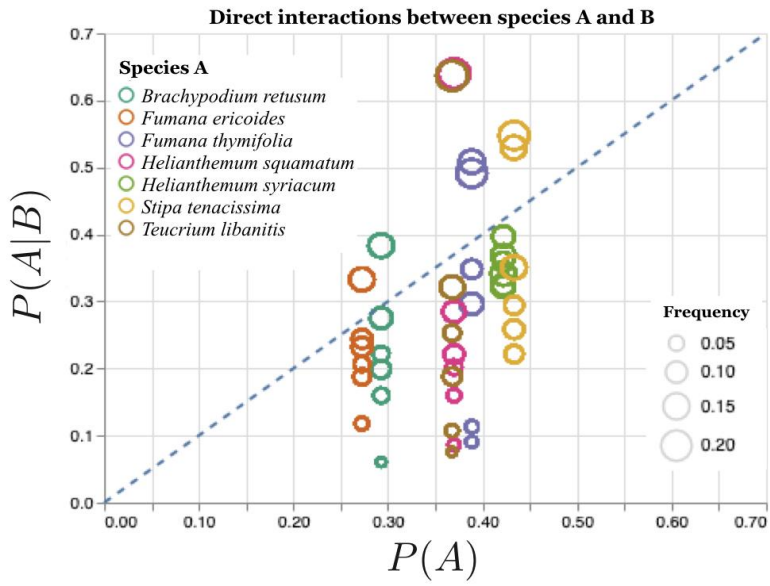

**b)**

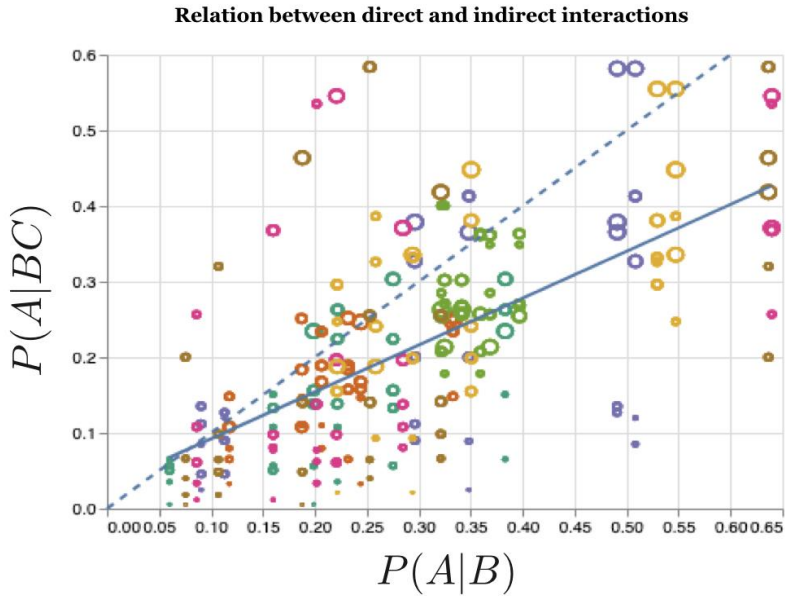

c)

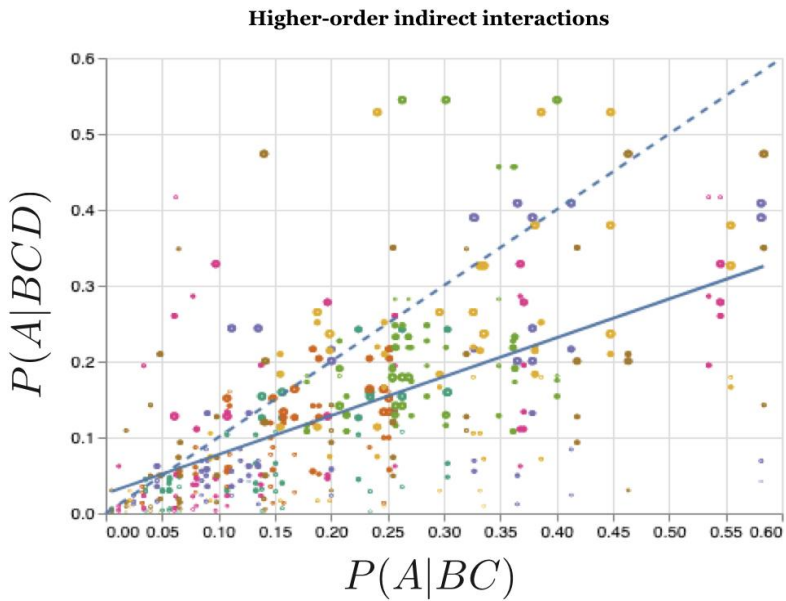

**Fig. 4.** The strength of direct and indirect species interactions. **a)** Relative abundance of species A P(A) vs the abundance of A when another species B is already present in the patch, P(A|B). **b)** Relation between the abundance of species A when B is present P(A|B) with the abundance of A when both B and C are present. The color coding in both plots corresponds to a fixed species A, while B, C and D are varied. **c)** Relation between triple (P(A|BC)) and higher-order interactions, represented by the conditional probability that species A is present when species B, C and D are already in the patch. In all the plots, the size of the circles indicates the relative abundance of a particular combination in the overall population. The dashed line is the diagonal x=y, which would correspond to the case of independent probabilities (no interactions). The solid blue line corresponds to a linear fit to the data.

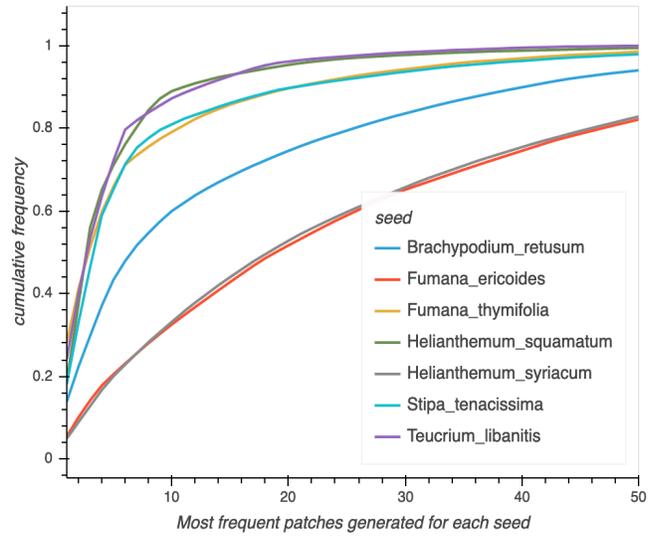

**Fig. 5** Cumulative distribution of unique patches generated by a single pioneer species introduced into the VAE's latent space.